\documentclass[letterpaper, 10 pt, journal, twoside]{IEEEtran} 
\usepackage{times,subfigure}
\usepackage{easyReview}
\usepackage{amsmath,amsfonts}
\usepackage{algorithmic}
\usepackage{mathtools}
\usepackage{algorithm}
\usepackage{array}
\usepackage{textcomp}
\usepackage{stfloats}
\usepackage{url}
\usepackage{verbatim}
\usepackage{graphicx}
\usepackage{xcolor}

\usepackage{cite}
\ifCLASSINFOpdf
\else
\fi

\begin{document}
\title{Unsupervised Meta-Testing with Conditional Neural Processes for Hybrid Meta-Reinforcement Learning}
\author{Suzan Ece Ada$^{1}$, and Emre Ugur$^{1}$%
\thanks{Manuscript received: March 28, 2024; Revised: June 30, 2024; Accepted: July 27, 2024.}
\thanks{This letter was recommended for publication by Editor Aleksandra Faust upon evaluation of the Associate Editor and reviewers' comments. This work has been supported by TÜBİTAK BİDEB 2224-A Grant Program for Participation in Scientific Meetings Abroad and European Union-funded INVERSE project (no. 101136067).} 
\thanks{$^{1}$Suzan Ece Ada and Emre Ugur are with Department of Computer Engineering, Bogazici University, Istanbul, Türkiye
        {\tt\footnotesize \{ece.ada,emre.ugur\}@bogazici.edu.tr}}%
}
\maketitle

\begin{abstract}
We introduce Unsupervised Meta-Testing with Conditional Neural Processes (UMCNP), a novel hybrid few-shot meta-reinforcement learning (meta-RL) method that uniquely combines, yet distinctly separates, parameterized policy gradient-based (PPG) and task inference-based few-shot meta-RL. Tailored for settings where the reward signal is missing during meta-testing, our method increases sample efficiency without requiring additional samples in meta-training. UMCNP leverages the efficiency and scalability of Conditional Neural Processes (CNPs) to reduce the number of online interactions required in meta-testing. During meta-training, samples previously collected through PPG meta-RL are efficiently reused for learning task inference in an offline manner. UMCNP infers the latent representation of the transition dynamics model from a single test task rollout with unknown parameters. This approach allows us to generate rollouts for self-adaptation by interacting with the learned dynamics model. We demonstrate our method can adapt to an unseen test task using significantly fewer samples during meta-testing than the baselines in 2D-Point Agent and continuous control meta-RL benchmarks, namely, cartpole with unknown angle sensor bias, walker agent with randomized dynamics parameters. 
\end{abstract}
\IEEEpeerreviewmaketitle
\begingroup
\renewcommand\thefootnote{}\footnote{%
© 2024 IEEE. Personal use of this material is permitted.  Permission from IEEE must be obtained for all other uses, in any current or future media, including reprinting/republishing this material for advertising or promotional purposes, creating new collective works, for resale or redistribution to servers or lists, or reuse of any copyrighted component of this work in other works. DOI: 10.1109/LRA.2024.3443496
}%
\addtocounter{footnote}{-1}
\endgroup
\vspace*{-4mm}
\section{Introduction}
\IEEEPARstart{D}{espite} the remarkable achievements of deep reinforcement learning (RL) algorithms \cite{schulman2017proximal,DBLP:journals/corr/MnihKSGAWR13}, they are hindered by poor sample efficiency and limited generalizability. Few-shot meta-reinforcement learning (meta-RL) framework\cite{beck2023survey,finn2017model} aims to overcome these limitations by learning a distribution of tasks during meta-training and adapting to a new task from the same distribution using a few samples from the new task in meta-testing. Given the high cost associated with collecting data in the real-world, minimizing the number of online interactions for rapid adaptation is essential for real-world applications.

Fast adaptation under uncertainty has various applications in the real-world, such as autonomous driving, disassembly with manipulation robots, medical diagnosis, and ventilation robots. Because the real-world is non-stationary, an agent should be robust to changing environmental dynamics, such as failures in the sensors and actuators \cite{https://doi.org/10.48550/arxiv.2203.01387}. Relying on deep RL agents trained on a single task can lead to failures in these settings. Thus, meta-RL \cite{finn2017model} emerges as a promising direction for increasing robustness through fast adaptation to an unknown test task. However, since a meta-RL algorithm requires samples from the test task for adaptation, it is limited by the number of samples collected from the unknown test task. Increasing sample efficiency reduces dependence on the costly and potentially dangerous process of collecting real-world experiences in an unknown task. This challenge is compounded when the reward signal from the test task is either unavailable or unreliable due to issues such as sensor malfunctions \cite{yang2019norml,beck2023survey}. To increase sample efficiency during meta-testing in settings with unknown rewards, we need to learn an expressive meta-model during training for cost-effective sample generation. 

This work investigates meta-RL settings where the transition function changes across a distribution of tasks and the reward signal is missing during meta-testing. Each transition function depends on a hidden dynamics parameter during meta-training and meta-testing. Imagine a setting where an agent is tasked with moving to a goal location in different environments parameterized by different force fields. Different force fields push the agent in different directions; requiring the agent to move robustly with different transition dynamics using few trial-and-error attempts. Similarly, in instances of a sensor malfunction, an unknown bias can result in inaccurate signals, necessitating the robot to adapt to this new environment with minimal damage \cite{clavera2018learning,yang2019norml}. In these cases, sample efficiency and not relying on reward signals are crucial, especially when engaging with a novel scenario, referred to as the ``test task'' where the number of real-world interactions is limited. No-reward Meta Learning \cite{yang2019norml} is an unsupervised meta-testing algorithm that adapts without reward signals using a learned advantage function. Similarly, Context-aware Dynamics Model (CaDM) \cite{lee2020context} and Rapid Motor Adaptation (RMA) \cite{kumar2021rss_rma} adapts within an episode without relying on rewards during meta-testing, though they use fixed length of transitions during training and testing.

We introduce a novel hybrid meta-RL formulation that decouples task inference from parameterized policy gradient-based (PPG) meta-RL\cite{finn2017model,beck2023survey}. Our approach diverges from prior works \cite{rakelly2019efficient,varibad} by targeting meta-testing sample efficiency through a modular approach while not requiring reward information during meta-testing. This strategy substantially improves sample efficiency in meta-testing by predicting next states given target states and actions queries, thereby generating cost-effective samples from the meta-model rather than relying on costly online environment interactions. Furthermore, Unsupervised Meta-Testing with Conditional Neural Processes (UMCNP) allows reusing samples collected from PPG meta-RL meta-training for task inference in an offline manner, without compromising sample efficiency during meta-training. This decoupling enhances the flexibility and applicability of our approach in scenarios where retraining the meta-policy with online interactions to improve meta-testing sample efficiency is not feasible. 

The key contributions of our method, UMCNP, include (1) improving sample efficiency via fast adaptation to the unseen test task (requires only a single rollout from the unseen test task compared to 25 rollouts used in prior works \cite{yang2019norml}), (2) generating and learning meta-dynamics models with no access to test task parameters and rewards while being permutation invariant and agnostic to the number of samples, (3) utilizing offline datasets for task inference in meta-RL with unsupervised meta-testing, and (4) decoupling PPG-based and task inference-based meta-RL, offering flexibility of application and enhancing informative trajectory generation. We evaluate our method using a point agent, a cartpole with sensor bias, and the widely used Walker Agent Environment featuring variable dynamics. Our results demonstrate a remarkable improvement in sample efficiency over the unsupervised meta-testing baselines.
\vspace*{-3mm}
\section{Related Work}
Meta-learning provides a framework for learning new tasks more efficiently from limited experience by using knowledge from previously learned tasks \cite{Schmidhuber:87long,finn2017model}. Meta-learning has demonstrated promising results in regression, classification, and deep RL \cite{finn2017model,yang2019norml,rothfuss2018promp,liu2022a,beck2023survey}. Our work builds on the meta-RL framework, where a distribution of tasks shares a common structure but differs in the transition dynamics model. The assumption of a common structure allows unsupervised meta-testing in the new task with unknown transition dynamics model parameters. 
\vspace*{-3mm}
\subsection{Black Box Few-Shot Meta-RL}
Black box few-shot meta-RL methods assume less structure on the meta-learning algorithm compared to parameterized policy gradient few-shot meta-RL methods while sacrificing generalization \cite{beck2023survey}. Prior work on black box meta-RL learns the task-specific context variable from a set of experiences collected by a control policy. Notably, Mishra et al. \cite{conf/iclr/MishraR0A18}, use memory-augmented models like temporal convolutions and soft attention to learn a contextual representation. Within the domain of meta-imitation learning, Duan et al. \cite{conf/nips/DuanASHSSAZ17} encode expert demonstrations into a learned context embedding where policy is conditioned on. 
\subsection{Parameterized Policy Gradient Based Few-Shot Meta-RL}
Parameterized policy gradient-based few-shot meta-RL methods \cite{gupta2018meta,li2017meta,finn2017model,beck2023survey,vuorio2019multimodal}, utilize policy gradient methods in the inner loop \cite{beck2023survey}. For instance, in Model-Agnostic Meta-Learning (MAML) \cite{finn2017model}, a parameterized policy gradient meta-learning algorithm is used where train and test tasks are sampled from the same distribution. No-reward Meta-Learning (NORML), proposed by Yang et al. \cite{yang2019norml}, is an extension of MAML-RL framework \cite{finn2017model} for unsupervised meta-testing where the environment dynamics model changes across tasks instead of the reward function. NORML aims to utilize past experience to quickly adapt to tasks with unknown parameters from a few samples without reward signals. While these methods have demonstrated promising results, they fall short in sample efficiency in meta-testing. We instead leverage the conditional neural processes \cite{cnp} to reduce the number of experiences an agent needs for adaptation to the test task.  

\subsection{Few-Shot Meta Reinforcement Learning with Task Inference}

Few-shot meta-RL methods that use task-inference attempt to infer the underlying MDP of the unknown test task in the meta-RL inner loop \cite{beck2023survey}. Zintgraf et al. \cite{varibad} uses a recurrent neural network (RNN) architecture to encode posterior over ordered transitions. The posterior belief is used to augment the state variable on which the policy is conditioned. Unlike FOCAL \cite{li2020focal}, which employs a deterministic encoder, PEARL \cite{rakelly2019efficient}, and VARIBAD \cite{varibad} use a stochastic encoder. Latter methods train the inference model by optimizing the evidence lower bound (ELBO)\cite{vae} objective. FOCAL, on the other hand, clusters transitions in the latent space using an inverse power loss distance metric learning objective under the assumption that tasks are deterministic and there exists a one-to-one function from the transitions to the $T \times R$ where $T$ and $R$ denote the task-specific transition function and reward function, respectively. In contrast to these approaches, while the goal of UMCNP is to identify the latent task parameter in the inner loop, we target few-shot meta-RL with unsupervised meta-testing where the reward information is missing. Moreover, our method decouples task inference from PPG in meta-RL, allowing flexibility. This separation improves sample efficiency in meta-testing even after training has concluded, as long as the samples from meta-training are accessible.

Rapid Motor Adaptation (RMA) \cite{kumar2021rss_rma}, categorized as a model-free task inference meta-RL approach \cite{beck2023survey}, have shown remarkable performance in sim-to-real applications on an A1 robot. RMA initially trains a base policy that takes the latent extrinsics vector encoded by the environment factor encoder, the state, and the previous action. Then, it trains an adaptation module to predict latent extrinsics from a history of state-action tuples. During deployment it uses the adaptation module and the base policy since the environment factors are hidden during testing. In contrast to the previous methods, RMA can adapt to the test task at test time within an episode. Although UMCNP can also adapt within an episode due to the flexibility to the number of samples used in training, it does not require environment parameters during training, which may not be available in a robotics dataset, and is only trained with transitions.

Model based meta-RL algorithms focus on learning the dynamics and the reward function of the environment. While they are sample efficient and enjoy the advantages of supervised learning, they can have lower asymptotic performance \cite{beck2023survey}. CaDM \cite{lee2020context} encodes a fixed number of transitions and, similar to RMA, can adapt within an episode. It can be combined with model-free RL by using the latent vector concatenated to the states to train a model-free policy gradient algorithm. CaDM trains the dynamics model and the context encoder by predicting forward and backward dynamics from a fixed number of state and action transitions. Unlike CaDM, UMCNP handles backward and forward state prediction in a permutation-invariant manner, agnostic to the number of observations, offering  flexibility during testing without the need to train for varying future and history lengths.

\subsection{Conditional Neural Processes}
Conditional Neural Processes (CNPs) \cite{cnp} predict the parameters of a probability distribution conditioned on a permutation invariant prior data and target input. CNPs aim to represent a family of functions using Bayesian Inference and the high representational capacity of neural networks \cite{BAO2024106201,bruinsma2023autoregressive,kim2018attentive,ye2022contrastive}. Prior approaches have leveraged the robust latent representations obtained by the CNP model in high-dimensional trajectory generation \cite{cnmp}, long-horizon multi-modal trajectory prediction \cite{cnmp,SEKER202222}, and sketch generation tasks \cite{GAN-CNMP}.

\section{Preliminaries}
\subsection{Notation}

We denote tasks as Markov Decision Processes (MDP). An MDP is represented by the tuple $(\mathcal { S },\mathcal{ A }, P, r, \rho_0,\gamma)$, where $\mathcal{S}$, and $\mathcal{ A } $ are the state and action spaces, $\rho_0(s_0)$ is the initial state distribution, $r$ is the reward function, and $ \gamma \in [0,1) $ is the discount factor \cite{sutton2018reinforcement}. $P{(\boldsymbol{s_{t+1}}|\boldsymbol{s_t}, \boldsymbol{a_t})}$ is the state transition dynamics where $(\boldsymbol{s_t}, \boldsymbol{a_t})$ are the state and action at timestep $\boldsymbol{t}$, and $\boldsymbol{s_{t+1}}$ is the next state. 

The objective of an RL agent is to maximize the expected cumulative discounted reward $\mathbb{E}_{\pi}\left[\sum_{t} \gamma^t r\left(\boldsymbol{s}_t, \boldsymbol{a}_t\right)\right]$ by learning a policy $\pi_\theta(\boldsymbol{a_t}|\boldsymbol{s_t})$, parameterized by $\theta$. 
Advantage function of a state and action tuple $A^{\pi}(s_t, a_t)$ can be formulated as $A^{\pi}(s_t, a_t) \coloneqq Q^{\pi}(s_t, a_t) - V^{\pi}(s_t)$ where Q-function and value function are $Q^{\pi}(s_t, a_t) \coloneqq \mathbb{E}_{s_{t+1:\infty}, a_{t+1:\infty}} \left[ \sum_{k=0}^{\infty} \gamma^k r_{t+k} \right]$, and $V^{\pi}(s_t) \coloneqq \mathbb{E}_{s_{t+1:\infty}, a_{t:\infty}} \left[ \sum_{k=0}^{\infty} \gamma^k r_{t+k} \right]$, respectively \cite{schulman2015high}. Advantage function estimates the relative advantage of taking an action in that state with respect to the estimated value of that state \cite{Ada_Ugur_Akin_2022}.

\subsection{Model-Agnostic Meta-Learning}

In meta-RL, we learn to adapt to a distribution of tasks represented by $p(T)$, where each task is distinguished by a unique Markov Decision Process (MDP) despite sharing a common structure. Typically, the tasks share identical state and action spaces, yet they vary in terms of their reward functions and state transition dynamics. A prominent approach for gradient-based meta-RL is the Model-Agnostic Meta-Learning (MAML) approach \cite{finn2017model}. During meta-training, MAML trains a meta-policy $\pi_\theta$, also known as an initial or a base policy \cite{beck2023survey}, characterized by parameters $\theta$. To adapt quickly to a new task from this distribution, the meta-policy requires only a small number of samples, $\mathcal{D}^{tr}_i$ following the equation

\begin{equation}
\phi_i = \theta + \alpha \nabla_{\theta} \mathcal{\hat{J}}(\theta, \mathcal{D}^{tr}_i)
\label{eq:mamladapt}
\end{equation}

where $\alpha$ is the adaptation learning rate, $\phi_i$ is the adapted policy parameter, and $\mathcal{\hat{J}}(\theta, \mathcal{D}^{tr}_i)$ is the estimated RL objective for task $T_i$ \cite{finn2017model}. After applying this adaptation step separately to a batch of tasks in the inner loop, we obtain $ \mathcal{D}^{test}_i$ from each task with the adapted parameters. Subsequently, we compute the meta-objective in the outer loop $\max_{\theta} \sum_{T_i \sim p(\mathcal{T})} \mathcal{\hat{J}} (\phi_i, \mathcal{D}^{test}_i)$:

\begin{equation}
\max_{\theta} \sum_{T_i \sim p(\mathcal{T})} \mathcal{\hat{J}} (\theta + \alpha \nabla_{\theta} \mathcal{\hat{J}}(\theta, \mathcal{D}^{tr}_i), \mathcal{D}^{test}_i)
\end{equation}

with respect to the meta policy parameters $\theta$ \cite{finn2017model,beck2023survey}. Once the meta-training is finished, the agent collects a few samples in the new test task during meta-testing and adapts its meta policy using Eq. \ref{eq:mamladapt}. 

\subsection{No-Reward Meta Learning}
Provided that the change in dynamics can be represented in a set of tuples of state ($\boldsymbol{s}$), action ($\boldsymbol{a}$), and next-state ($\boldsymbol{s'}$) from the same task $\{\left(\boldsymbol{s}, \boldsymbol{a}, \boldsymbol{s'}\right)\}$, NORML learns a pseudo-advantage function $A_{\psi}\left(\boldsymbol{s}, \boldsymbol{a}, \boldsymbol{s'}\right)$ \cite{yang2019norml}. It is important to note that the aim of $A_{\psi}$ is to guide the meta-policy adaptation in Eq. \ref{eq:mamladapt} instead of fitting to the advantage function, thus we refer to the learned advantage function as the pseudo-advantage function. Task-specific gradient update in NORML, $\boldsymbol{\phi}_{i}=\boldsymbol{\theta}+\boldsymbol{\theta}_{\text {offset}}+\alpha \nabla_{\boldsymbol{\theta}} \mathcal{\hat{J}}_{\mathcal{T}_{i}}^{\text {NoRML }}\left(\boldsymbol{\theta}, D_{i}^{\text {train }}\right)$, is computed \cite{yang2019norml} with 

\begin{equation}
\boldsymbol{\theta}+\boldsymbol{\theta}_{\text {offset}}+\alpha \sum_{D_{i}^{\text {train }}} A_{\boldsymbol{\psi}}\left(\boldsymbol{s}, \boldsymbol{a}, \boldsymbol{s}'\right) \nabla_{\boldsymbol{\theta}} \log \pi_{\boldsymbol{\theta}}\left(\boldsymbol{a} \mid \boldsymbol{s}\right)
\label{eq:normladapt}
\end{equation}

where $A_{\psi}$ is used to compute task-specific parameters in the meta-learning inner loop from a set of state, action, and next-state transitions of task $i$ denoted as $D_{i}^{\text {train }}$ that does not contain reward signals. In this adaptation phase, the pseudo-advantage function guides the policy gradient update. To encourage decoupling between the meta policy and the task-specific policy in the inner loop, we learn an offset parameter $\boldsymbol{\theta}_{\text {offset }}$ end-to-end \cite{yang2019norml}. This offset parameter, added in the inner loop adaptation formulated in Eq. \ref{eq:normladapt}, is identical for all tasks. 

The learned pseudo-advantage function $A_{\psi}$ is optimized in the meta-learning outer loop. Using $D_i^{\text {train }}$, and $D_i^{\text {test}}$, we not only update $\boldsymbol{\psi}$ and  $\boldsymbol{\theta}$ in the outer loop, but additionally update the offset parameters $\boldsymbol{\theta}_{\text {offset}}$ \cite{yang2019norml}:

\begin{equation}
\begin{bmatrix}
\theta \\
\theta_{\text{offset}} \\
\psi
\end{bmatrix}
\leftarrow
\begin{bmatrix}
\theta \\
\theta_{\text{offset}} \\
\psi
\end{bmatrix}
+ \beta \sum_{T_i \sim \mathcal{T}} \nabla_{\left[\theta^T \theta^T_{\text{offset}} \psi^T\right]^T}
\mathcal{\hat{J}}_{T_i} (\phi_{i}, D^{\text{test}}_i )
\label{eq:normlouterloop}
\end{equation}

where  $\nabla \mathcal{\hat{J}}_{T_i}(\phi_{i}, D^{\text{test}}_i)$ is the gradient of the meta objective formulated as $\sum_{D^{\text{test}}_i} A^\pi(s, a) \nabla \log \pi_{\phi_{i}}(a | s)$, $A^\pi(s , a)$ is the  RL advantage function, and $\beta$ is the learning rate. 

\subsection{Conditional Neural Processes (CNP)}

The architecture of CNP consists of parameter-sharing encoders and a decoder named the query network \cite{cnp}. Initially, a set of random input and output pairs ${(x_{f^{i}},y^{true}_{f^{i}})}$ are sampled from a function ${f^{i}}\in F$. These pairs are then encoded into latent representations through parameter-sharing encoder networks. Subsequently, the average of these representations is computed, further ensuring that the model is invariant to permutations and the number of inputs. The averaged representation is then concatenated with the target input query and passed to the query network. Finally, the query network generates the predicted mean and standard deviation for the queried input $(x_{f^{i}}^q)$. 

\begin{figure*}
	\begin{minipage}{0.99\columnwidth}
		\includegraphics[width=\textwidth]{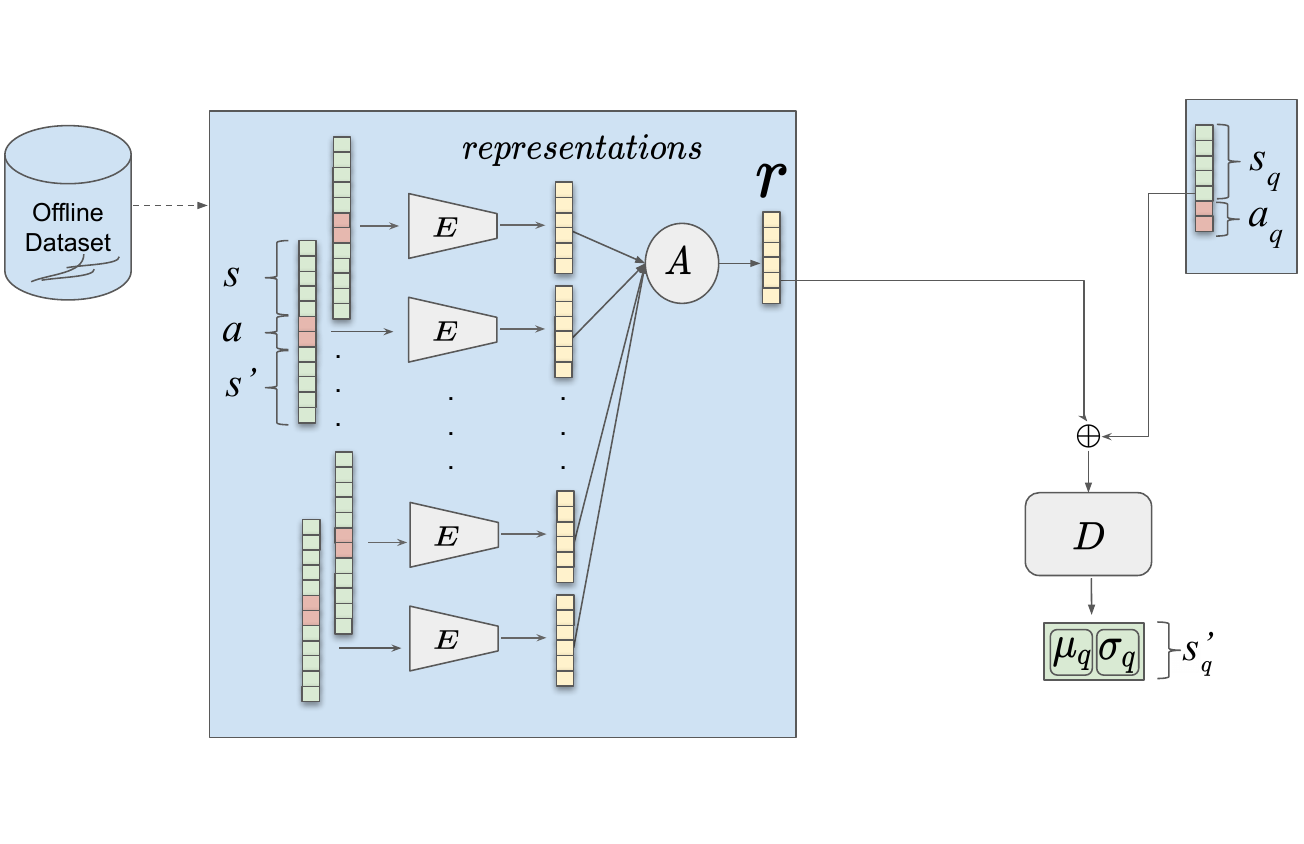}
		\caption{Meta-training procedure of our proposed method UMCNP. Randomly selected samples from a task are extracted from the offline dataset and encoded into fixed-size representations. An averaging module $A$ computes the mean of these representations, denoted by $\mathbf{r}$. This mean representation, $\mathbf{r}$, is then concatenated with the state query, and action query from the same task $[s_q, a_q]$, also drawn from the offline dataset, are fed to the decoder network as $[s_q, a_q, r]$. The decoder then outputs the mean $\mu_q$ and standard deviation $\sigma_q$ for the next state $\mathbf{s'}$ conditioned on the $[s_q, a_q, r]$ input vector.}
		\label{fig:umcnp}
    \vspace*{-2mm}
	\end{minipage} \hspace{2mm}
	\begin{minipage}{0.99\columnwidth}
   \vspace*{1mm}
		\includegraphics[width=\textwidth]{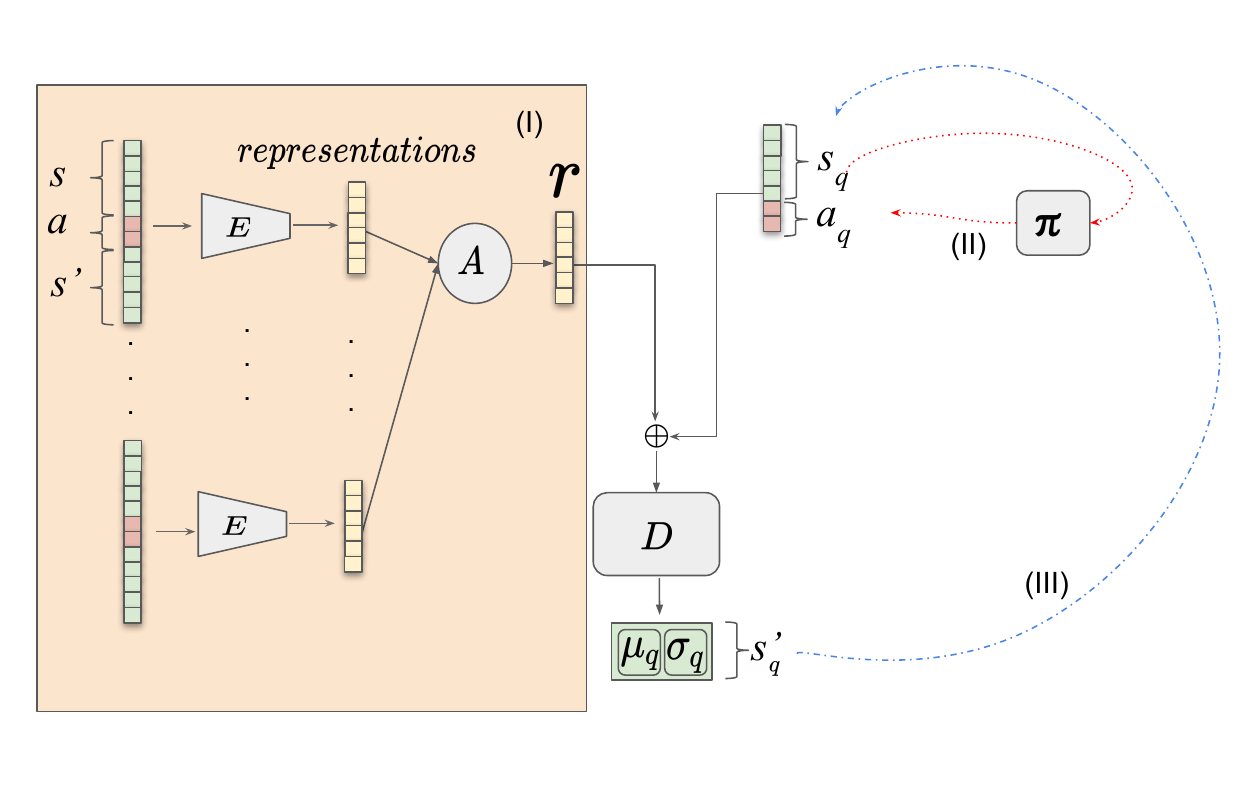}
		\caption{Meta-testing procedure of our proposed method UMCNP. The meta-policy is allowed to collect a few samples from the test task. (I) These samples ($\{\left(s_{k}, a_{k},s'_{k}\right)\}$) are encoded into fixed-size representations and averaged using module $A$. (II) Then, the meta-policy samples the action query $a_q$ conditioned on the initial state query $s_q$ to obtain $[s_q, a_q]$, which is concatenated with the mean representation $\mathbf{r}$ before being input to the decoder network. The decoder's predicted next state is used to sample the meta-policy's next action query. Through this method, we sequentially construct a complete trajectory, leveraging the dynamic modeling capabilities of our model.}
		\label{fig:umcnptest}
  \vspace*{-2mm}
	\end{minipage}
\end{figure*}

\section{Proposed Method: Unsupervised Meta-Testing with Conditional Neural Processes (UMCNP)}

In this section, we present our method, Unsupervised Meta-Testing with Conditional Neural Processes (UMCNP), for learning dynamics models by leveraging meta-training trajectories collected by PPG meta-RL. By sampling actions from the policy, the UMCNP model facilitates the autoregressive generation of complete trajectories. This process involves querying the model with the action and the state, receiving a prediction of the next state, and then feeding this state back into the model alongside new actions. A learned pseudo-advantage function takes state, action, and next state as input and guides the policy gradient in the adaptation. As a result, UMCNP effectively creates models of unknown environments with fewer samples, a significant advancement we detail after outlining the problem context.

Our method aims to enhance sample efficiency in meta-RL with unsupervised meta-testing. UMCNP requires fewer samples from the test task to generate trajectories without accessing environment parameters or rewards, addressing the challenge of unsupervised meta-testing. This procedure facilitates the generation of cost-effective trajectories for update during meta-testing.

\paragraph{Gradient-Based Meta Reinforcement Learning with Unsupervised Meta-Testing}
During meta-training, we store transitions for each $task_i$ as a set of transition tuples $B_i=\left\{\left(s, a, s'\right)\right\}_{t=0}^{H-1} \subset S \times A \times S$ without the task parameter and the rewards. We then use these batches to create our offline dataset illustrated in Fig. \ref{fig:umcnp}.

Both $A^\pi(s, a)$ and the pseudo-advantage function use polynomial regression with a fitted value function \cite{yang2019norml}. Policy parameters in the inner loop adaptation in Eq. \ref{eq:normladapt} and the outer loop meta-optimization in Eq. \ref{eq:normlouterloop} are updated using vanilla PG and actor-critic Proximal Policy Optimization \cite{schulman2017proximal} method with a fitted value function, respectively. We use Adam optimizer \cite{kingma2014adam} for both the inner loop and the \cite{yang2019norml} outer loop update. After we learn the meta-policy, offset parameter $\mathbf{\theta}_{offset}$ and the pseudo-advantage function $A_{\psi}\left(\boldsymbol{s}, \boldsymbol{a}, \boldsymbol{s'}\right)$ using NORML, we train our UMCNP model using the offline dataset of transitions $D= \{B_{i}\}_{i=1}^{n}$ where n denotes the number of transitions. 

\paragraph{Unsupervised Meta-Testing with Conditional Neural Processes (UMCNP)}

We train the UMCNP offline using unlabeled batches of transition tuples we previously collected during the first phase of online PPG meta-training. Fig.  \ref{fig:umcnp} illustrates the second phase of the meta-training procedure for UMCNP directed towards task inference. Notably, the transition dynamics parameter is hidden, and the data does not contain reward information during training and testing. In each meta-training iteration, we randomly select an unlabeled batch $B_i=\left\{\left(s, a, s'\right)\right\}_{t=0}^{H-1}$ from the offline dataset $D$. From this batch, we randomly select a set of task-specific transition tuples $\left\{\left(s_{k}, a_{k},s'_{k}\right)\right\}_{k}$ and a target query tuple $(s_q, a_q,s'_q)$ without replacement. We utilize $(s_q, a_q,s'_q)$ for the target state-action query $[s_q, a_q]$ and true target next state label $[s'_q]$.  

Next, we encode each tuple $\left(s_{k}, a_{k},s'_{k}\right)$ into a fixed size representation using a parameter sharing encoder network. These representations are then passed through an averaging module $A$ to obtain a learned latent vector $r$ that represents the hidden environment transition function parameter used in batch $B_i$. We then concatenate the resulting latent representation $r$ with the $[s_q, a_q]$ to predict the distribution parameters $\mu_q, \sigma_q$ of the next state $s'_q$ conditioned on $r$ and the target query $[s_q, a_q]$:

\begin{equation}
\mu_q, \sigma_q = f_{\theta_{D}}\left([s_q, a_q] \oplus \frac{1}{k}\sum_{k} g_{\theta_{E}}\left(s_{k}, a_{k}, s'_{k}\right)\right)
\label{eq:umcnpoutput}
\end{equation}

where $f_{\theta_{D}}$, $g_{\theta_{E}}$ are the decoder and the encoder networks, $\left(s_{k}, a_{k}, s'_{k}\right)$ are the randomly sampled transitions from the set $B_i$. The loss function of UMCNP can be expressed as

\begin{equation}
\mathcal{L}(\theta_{E}, \theta_{D})= -\log P\left(s_{q}^{'true} \mid \mu_q, \textit{softplus}(\sigma_q)\right).
\label{eq:umcnploss}
\end{equation}

The softplus activation function is applied to $\sigma_q$, and the resulting output, along with $\mu_q$, are used to compute a multivariate normal distribution with a diagonal covariance. A high variance in this distribution can indicate high uncertainty in the predicted next state value.

During meta-testing in Fig.  \ref{fig:umcnptest}, an agent collects a few samples $\{\left(s_{k}, a_{k},s'_{k}\right)\}$ from the unseen test task without rewards. A trained encoder network with shared weights encodes each sample into a fixed-size representation. Fig. \ref{fig:umcnptest} (I) shows how an averaging module ($A$) forms a learned shared representation of the test task dynamics. Using this shared latent vector representing the hidden task parameter, we can generate cost-effective rollouts using our meta-policy. This representation helps in predicting the parameters of the next-state distribution with the state-action query $[s_q, a_q]$. 

For every rollout generated by UMCNP, we use the same true initial state sampled from the true test task. We input this initial state into the stochastic meta-policy to obtain the action query $a_q$ depicted in Fig. \ref{fig:umcnptest} (II). Due to the stochasticity of the meta-policy trained with PPO, we can generate diverse rollouts from our UMCNP model which can facilitate informative exploration for our meta-policy. We then take the predicted next state to determine the subsequent action query in Fig. \ref{fig:umcnptest} (III), repeating the process of (III-II) and querying the decoder until the episode concludes. We combine these rollouts generated from UMCNP with the ground-truth rollout sampled from the test task. 

The learned pseudo-advantage network takes a tuple $(s,a,s')$ as input and outputs an advantage estimation value. Hence, the pseudo-advantage network $A_\psi(s,a,s')$ computes advantage estimates for each tuple in the combined set of generated rollouts and a ground-truth rollout. Finally, we perform a gradient update on the meta-policy for fast adaptation to the test task. This update uses the estimated advantage values, the learned parameter offset, and the ensemble of rollouts generated by UMCNP alongside a single rollout from the test task. Although the reward information is completely missing in meta-testing and only available during the first phase of meta-training, our model is capable of remarkably sample-efficient unsupervised adaptation.

\section{Experiments}
In this section, we analyze the performance of our method and compare it with meta-RL baselines in 2D point agent and continuous control tasks, including cartpole with sensor bias and biped locomotion of a walker agent with random dynamics parameters. ORACLE agent, trained using NORML, serves as a baseline to demonstrate the potential performance when extensive sampling from the test task is allowed. Unlike ORACLE \cite{yang2019norml}, UMCNP, NORML \cite{yang2019norml}, Probabilistic Ensembles with Trajectory Sampling CaDM (PE-TS CaDM), Vanilla CaDM\cite{lee2020context}, PE-TS \cite{chua2018deep}, Proximal Policy Optimization-CaDM (PPO-CaDM) \cite{lee2020context}, Vanilla DM \cite{lee2020context} have limited access to the test task during meta-testing. Despite this limitation, UMCNP achieves fast adaptation to the test task with significantly less interaction with the test task than NORML, matching ORACLE's performance. In all our experiments, we report results with 95\% confidence intervals with algorithms trained with 5 random seeds.
\begin{figure*}
\vspace*{0.5mm}
	\centering
	\begin{minipage}{0.88\textwidth}
		\centering
        \includegraphics[width=\textwidth]{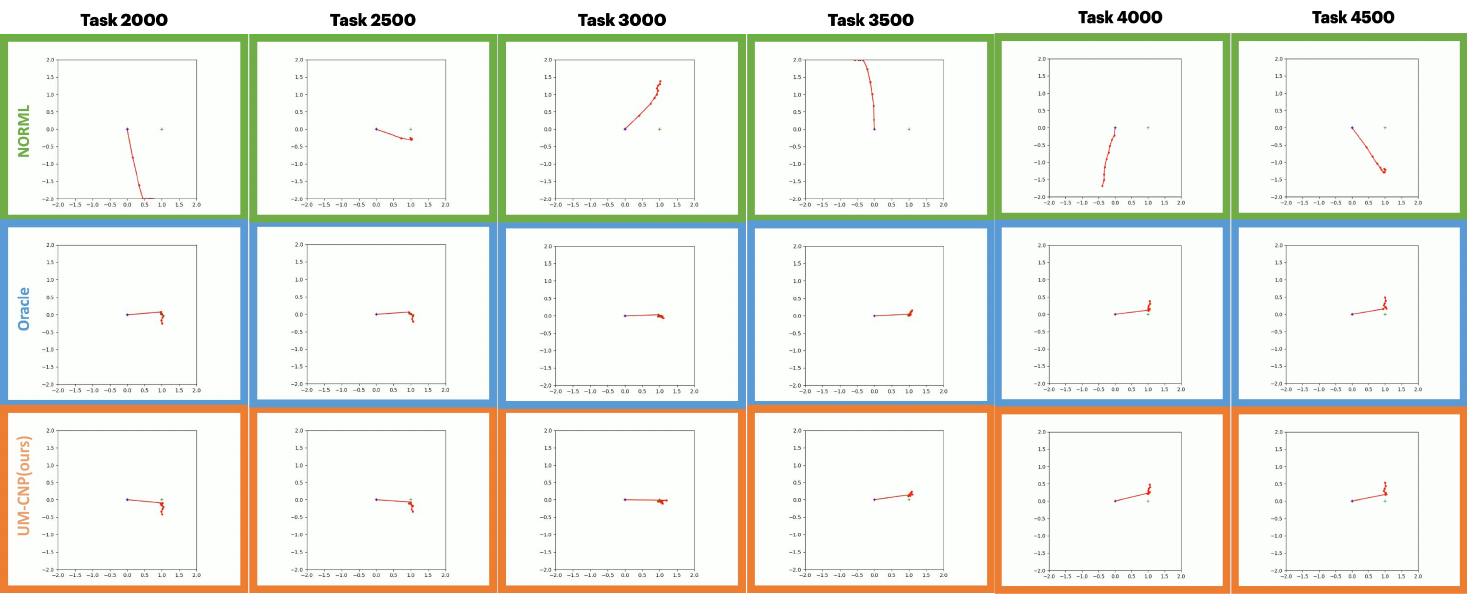}
		\caption{Test task indices 2000, 2500, 3000, 3500, 4000, and 4500 correspond to task parameter $\omega$ with values $-2\pi/10$, $0$, $2\pi/10$, $4\pi/10$, $6\pi/10$, $8\pi/10$ respectively.}
            \label{fig:task0}
	\end{minipage}%
    \vspace*{-5mm}
\end{figure*}

\begin{figure}
\includegraphics[width=\columnwidth]{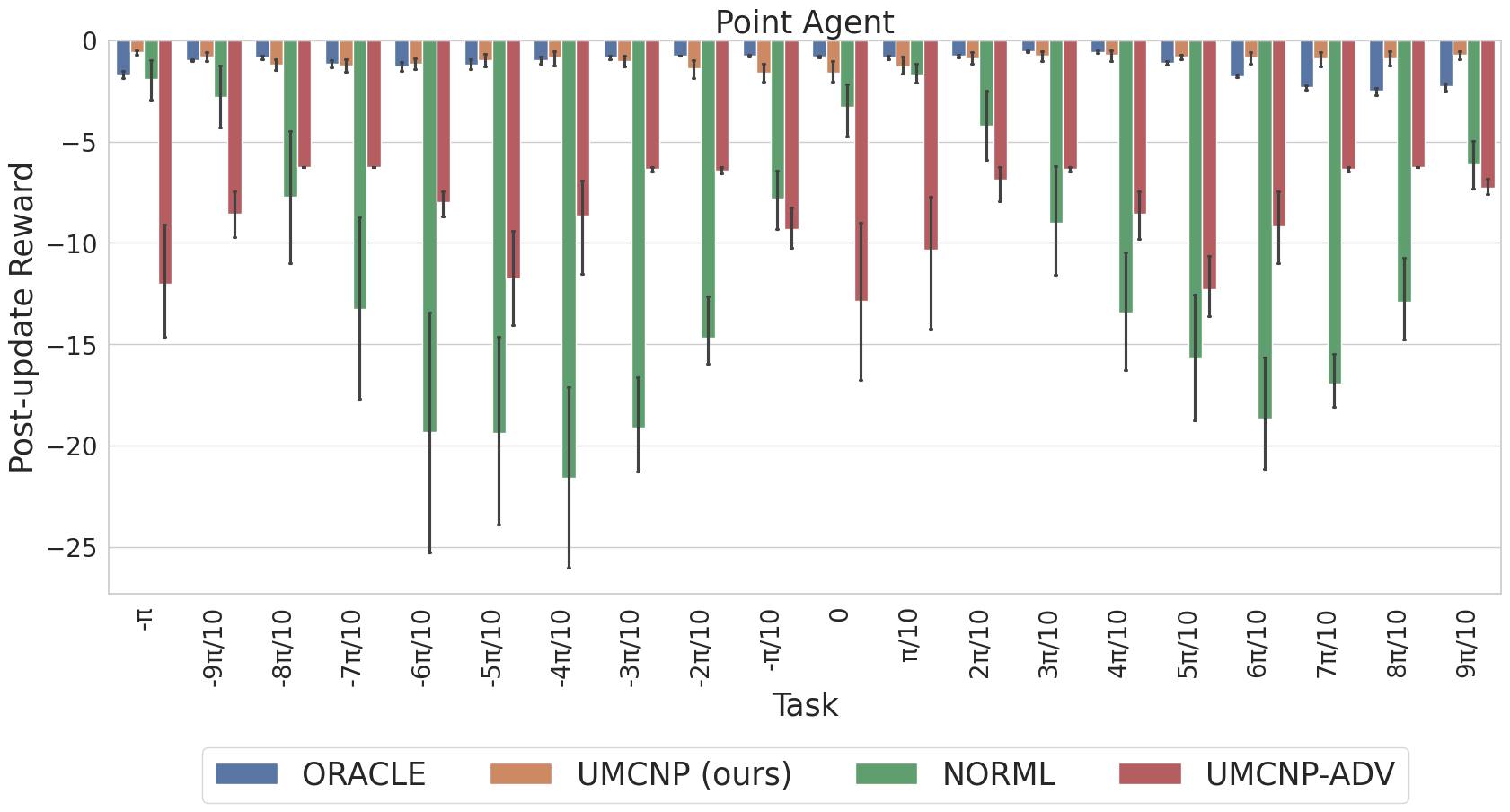}
\caption{Point agent plot shows the post-update cumulative reward (sum of rewards) (y-axis) in test tasks, computed after finetuning with 24 generated rollouts and 1 actual rollout for UMCNP (ours) (orange), 1 actual rollout for NORML \cite{yang2019norml} (green), and 25 actual rollouts for ORACLE/Ground-truth (blue) and UMCNP-ADV (red) with 95\% confidence intervals.}
\label{fig:metacnp}
\vspace*{-3mm}
\end{figure}

\subsection{2D Point Agent with Unknown Artificial Force Field}
The goal of the point agent, initialized at $(x=0,y=0)$, is to move to the position $(x=1,y=0)$, where $(x,y)\in [-2,2] \times [-2,2]$ are the positions on the 2D plane. We are interested in a meta-RL setting used in \cite{yang2019norml} where the reward function is identical across multiple tasks. Different tasks are created by generating different artificial force fields that push the agent in different directions using the parameter ($\omega$). Dynamics of the 2D point task are formulated as:

$$
\scriptstyle{
\left[\begin{array}{l}
s^{(1)}_{t+1} \\
s^{(2)}_{t+1}
\end{array}\right]=\left[\begin{array}{cc}
\cos \omega & -\sin \omega \\
\sin \omega & \cos \omega
\end{array}\right]\left[\begin{array}{l}
a^{(1)}_{t} \\
a^{(2)}_{t}
\end{array}\right]+\left[\begin{array}{l}
s^{(1)}_{t} \\
s^{(2)}_{t}
\end{array}\right]}
$$

where state vector denoted by $\left[s^{(1)}_{t},s^{(2)}_{t}\right]^T$ corresponds to the x, y-coordinates on the 2D plane, rotation parameter is defined by $\omega$, and the action vector at time $t$, $\left[a^{(1)}_{t},a^{(2)}_{t}\right]^T$, refer to the movement in the x,y-directions with $(d x_{t}, d y_{t})\in \left[-1,1\right]$ \cite{yang2019norml}. The maximum episode length in this environment is 10. We use the same reward function, the negative Euclidean distance from the goal position, and hyperparameters in NORML for comparative analysis. In the Point Agent environment, 5000 tasks are defined over the $[-\pi,\pi]$ interval. The agent is initially trained across a distribution of 5000 tasks, i.e. in environments with 5000 different force fields. Then, it is tested in unseen test tasks. The rollouts obtained from direct interactions with the test task will be referred to as actual rollouts. We have also trained our UMCNP model with the advantage values corresponding to the state, action, next state tuples recorded during meta-training and we name this variation as UMCNP-ADV. UMCNP-ADV takes in state, action, next state, and advantage tuples and predicts the mean and standard deviation of the next state and the advantage for the state, action target query. At meta-test time, the ORACLE agent is allowed to collect 25 actual rollouts from the unseen test task. Conversely, NORML agent, UMCNP and UMCNP-ADV are allowed to use a single actual rollout for fine-tuning the meta-policy. For a reliable evaluation, we use the same actual rollout data for UMCNP and UMCNP-ADV models that is used for the single rollout NORML fine-tuning. 
\begin{table}[!t]
\caption[Point Environment]{2D Point Agent Environment Post-update Reward with 95\% Confidence Intervals \label{table:pointtable}}
\centering 
\begin{tabular}{l||c|c}
\hline \hline $\mathbf{Point}$ $\mathbf{Agent}$ & $\mathbf{mean}$&$\mathbf{median}$\\
\hline \textbf{UMCNP(ours)}&$\textbf{-1.02}$ $\pm$ 0.09 & $\textbf{-0.91}$\\
\hline NORML & -11.49 $\pm$ 1.44 & -11.01 \\
\hline UMCNP-ADV &  -8.49 $\pm$ 0.58 & -7.46 \\
\hline ORACLE  & -1.20 $\pm$ 0.12 &-0.94 \\
\hline 
\end{tabular}
\vspace*{-3mm}
\end{table} 
Fig. \ref{fig:task0} illustrates the trajectories followed by the adapted policies where the trajectories of UMCNP and ORACLE agents are similar. Fig. \ref{fig:metacnp} shows the post-update cumulative reward obtained in the test tasks for ORACLE, UMCNP, NORML, and UMCNP-ADV with blue, orange, green, and red respectively. The results in Table \ref{table:pointtable} show the mean with 95\% confidence intervals and median post-update cumulative rewards collected in the test tasks. Adaptation using 24 generated rollouts from the UMCNP model achieves -1.02 $\pm$ 0.09, similar to the ground truth ORACLE with (25 actual rollouts) -1.20 $\pm$ 0.12, whereas NORML underperforms with -11.49 $\pm$ 1.44. Although UMCNP-ADV performs better than NORML, we observe that the UMCNP is significantly better than UMCNP-ADV. Future work will explore architectural improvements, such as separate decoders, to enhance UMCNP-ADV’s performance.

\subsection{Cartpole with Unknown Angle Sensor Bias}
\begin{figure}
\vspace*{1mm}
	\centering
		\includegraphics[width=0.43\columnwidth]{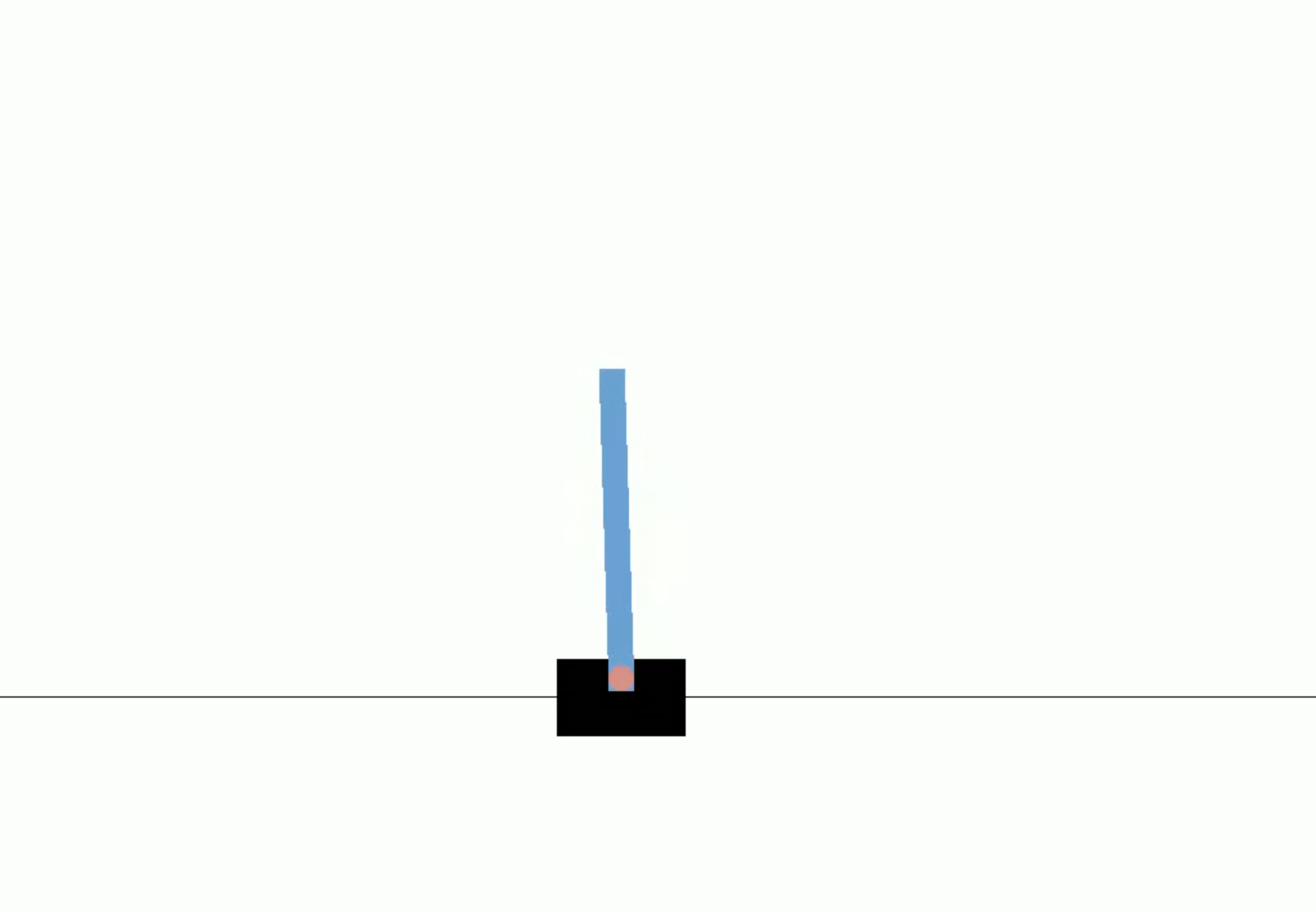}
		\includegraphics[width=.37\columnwidth]{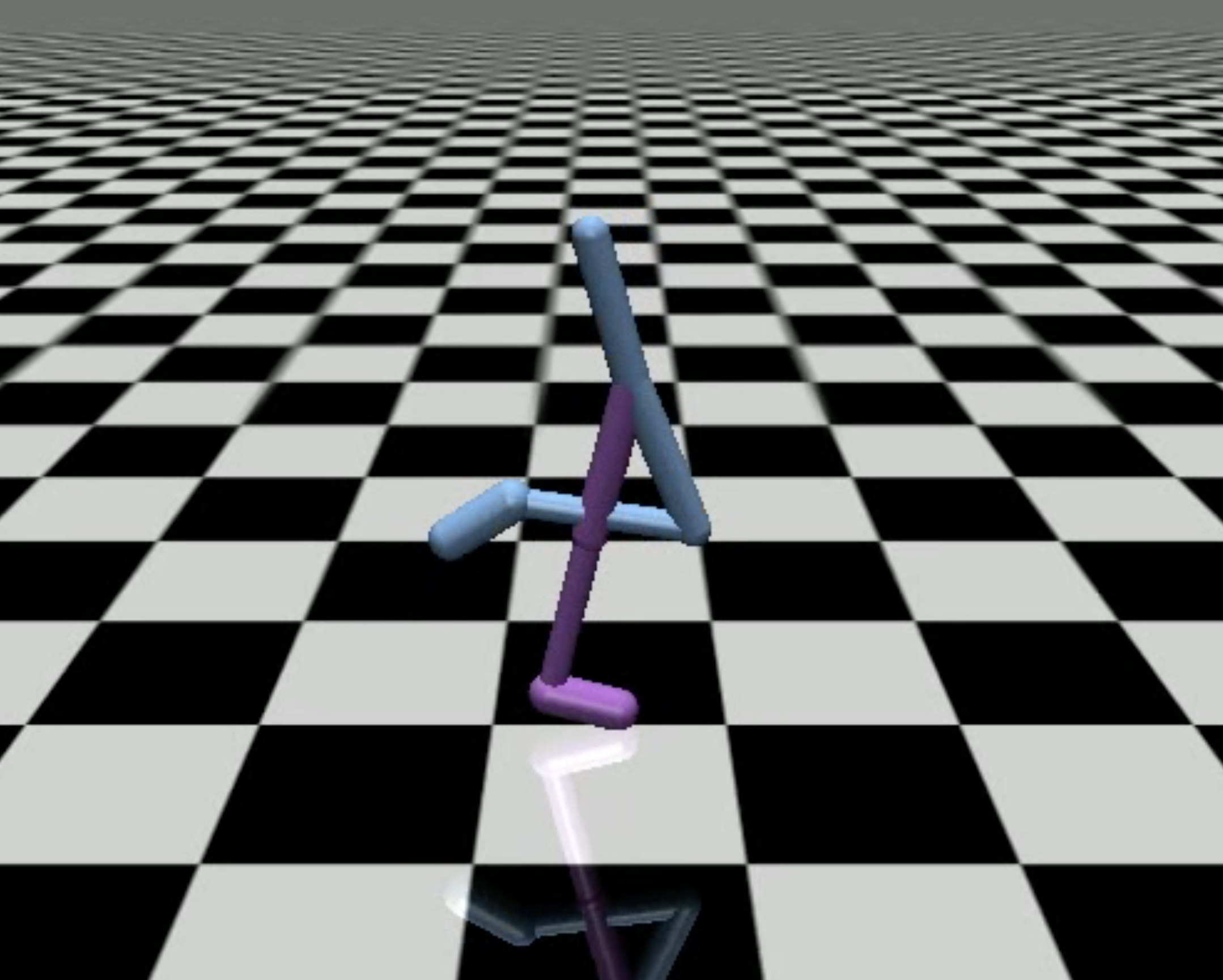}
	\caption{Visualization of Cartpole with Unknown Angle Sensor Bias Environment (left), Walker-2D Agent Environment with Random Dynamics Parameters (right) environments.}
	\label{fig:wenv}
    \vspace*{-2mm}
\end{figure}

\begin{figure}[htbp]
	\centering
		\subfigure[]{
		\includegraphics[width=.9\columnwidth]{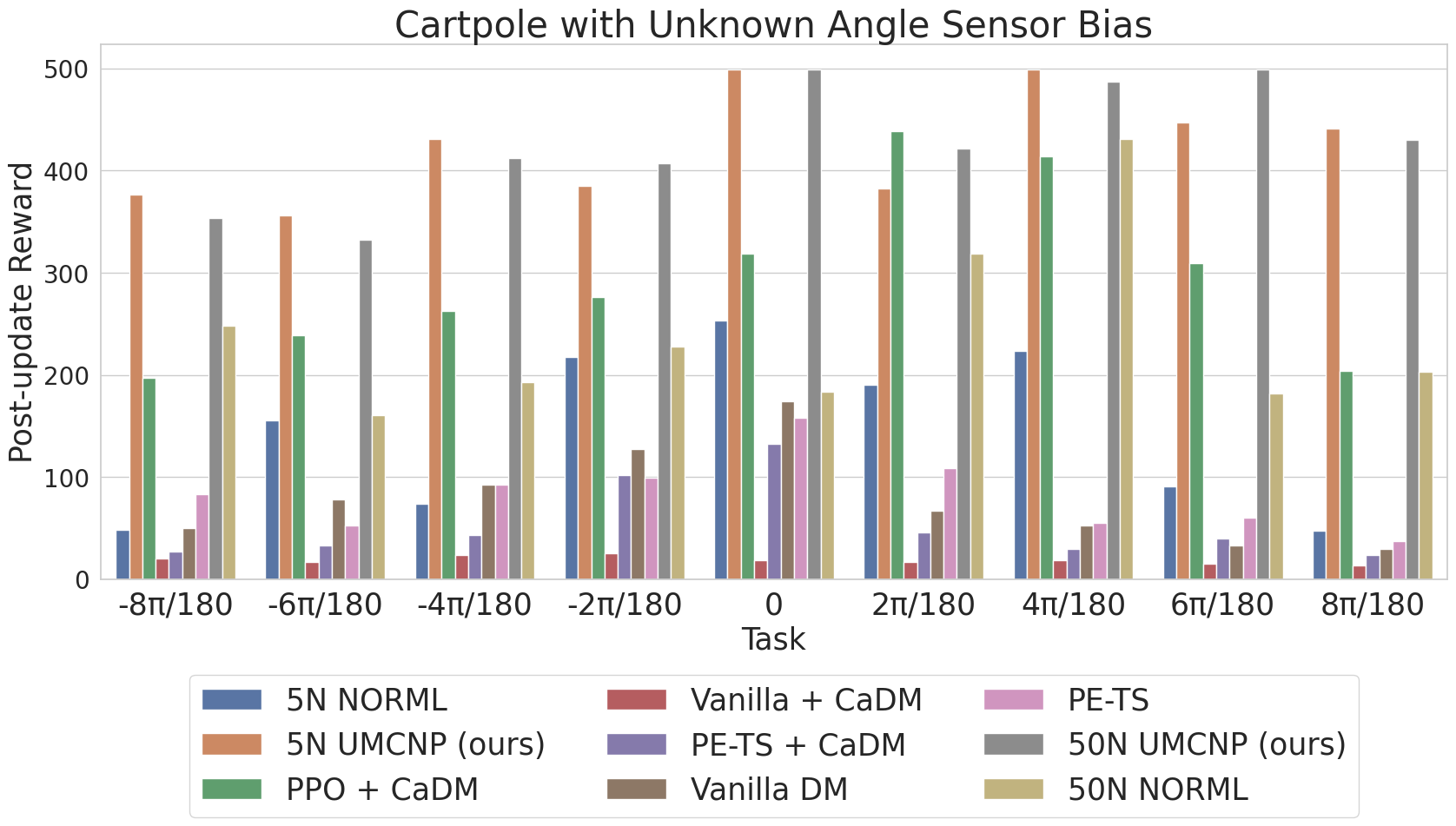}
	}
		\subfigure[]{
		\includegraphics[width=.9\columnwidth]{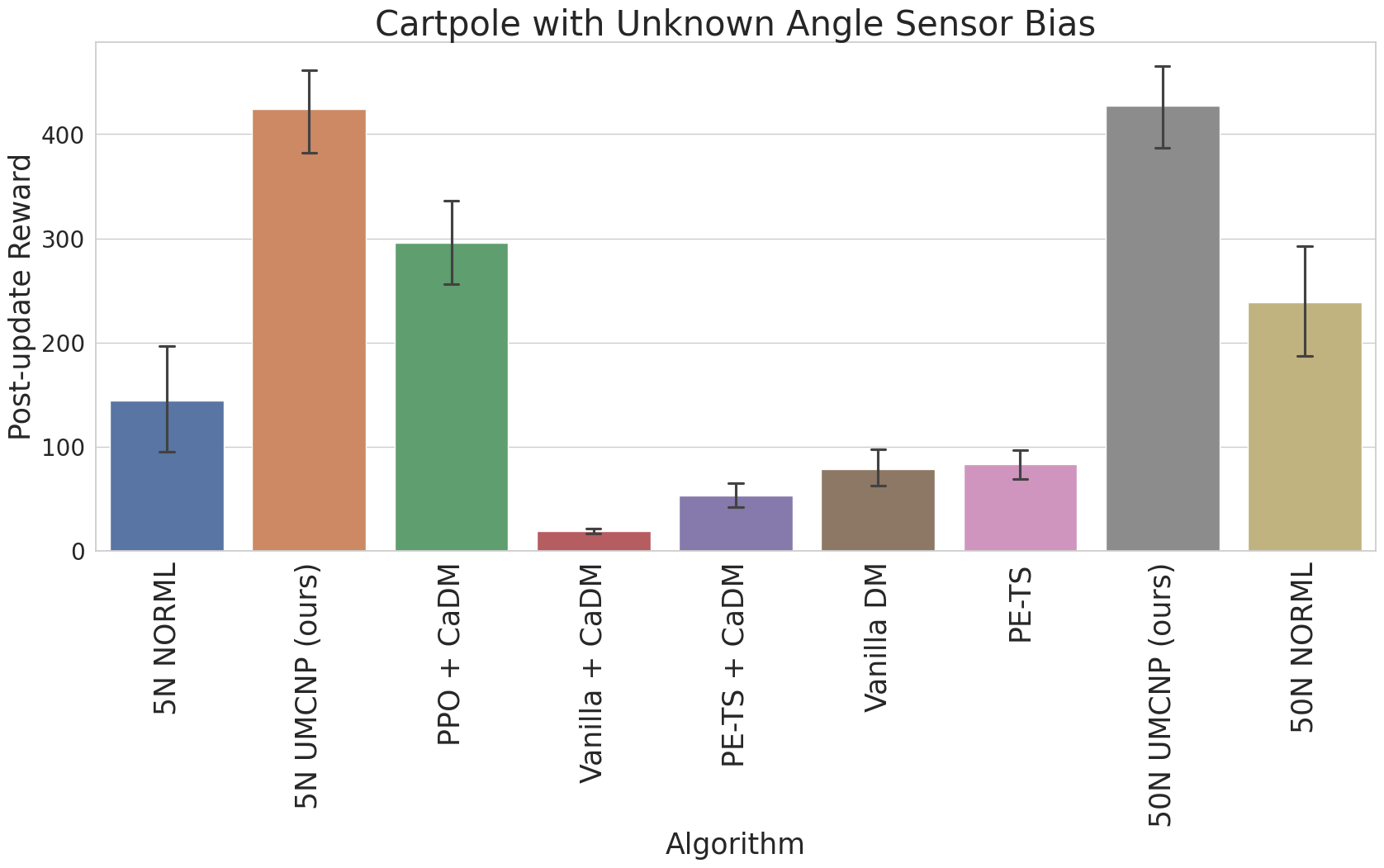}
	}
	\caption{(a) Post-update cumulative Reward (y-axis) in test tasks compared with baselines. (b) Overall performance compared with baselines with 95\% Confidence Intervals. 5N UMCNP (ours), 5N NORML and 50N UMCNP (ours), 50N NORML uses 5 and 50 test task transitions respectively. X-axis represents the test task parameter.}
	\label{fig:cartpoleexp}
     \vspace*{-4mm}
\end{figure}

We evaluate our method on the Cartpole with Angle Sensor Bias environment (Fig. \ref{fig:wenv}) \cite{yang2019norml}, an extension of the Cartpole environment simulated in MuJoCo, Gym \cite{todorov2012mujoco,brockman_gym_2016}. In this experiment, tasks differ by a hidden position sensor drift parameter in the range $[-8^{\circ},8^{\circ}]$. In meta-training, tasks are sampled from a uniform distribution where the hidden variable parameterizes the dynamics function. The state space consists of the position/angle and velocity/angular velocity of the cart and the pole. To evaluate the adaptation performance where a limited number of transition tuples are available when the maximum episode length of the environment is 500, we compared the performance of UMCNP to NORML with only 5 and 50 transitions, and PE-TS CaDM\cite{lee2020context}, Vanilla CaDM\cite{lee2020context}, PPO-CaDM \cite{lee2020context}, trained with history-length and future transition length of 5 and Random Shooting with 1000 candidates for planning following the Cartpole implementation details in \cite{lee2020context}, PE-TS \cite{chua2018deep}, Vanilla DM \cite{lee2020context} by using the code available in \cite{cadmcode}. Importantly, CaDM requires access to the reward function during training for Model Predictive Control \cite{garcia1989model} and we have provided the reward function definitions during training. Consistent with other experiments, results in Fig. \ref{fig:cartpoleexp} and Table \ref{tab:cartwalk} indicate that UMCNP outperforms the baselines and CaDM performs better when integrated in model-free RL.

\subsection{Walker-2D  with Random Dynamics Parameters}
In order to examine the sample efficiency of UMCNP in complex, higher dimensional control tasks, we evaluate our method in the Walker-2D Agent Environment with random dynamics parameters \cite{rothfuss2018promp} with a maximum episode length of 1000 in an unsupervised meta-testing setup where the agent receives no reward signal from the test task. Different locomotion tasks are created by unknown body mass, body inertia, damping on different joints, and friction parameters. The scaling parameters for body mass, body inertia, and friction parameters are sampled from a range determined by raising 1.5 to powers uniformly distributed between $-3$ and $3$, in the interval $[1.5^{-3}, 1.5^{3}]$. Likewise, the scaling parameters for damping across different joints follow the same methodology but use a base of 1.3 covering the range $[1.3^{-3}, 1.3^{3}]$. Walker environments with slimmer distributions, for instance  only parameterizing the body mass or reducing the interval of the uniform distributions does not show a significant increase in performance for NORML compared to domain randomization. These scenarios do not require sophisticated adaptation methods; hence, we use more established meta-RL benchmarks \cite{rothfuss2018promp,yang2019norml}.

\begin{figure}[htbp]
  \vspace*{-2mm}
\includegraphics[width=1\columnwidth]{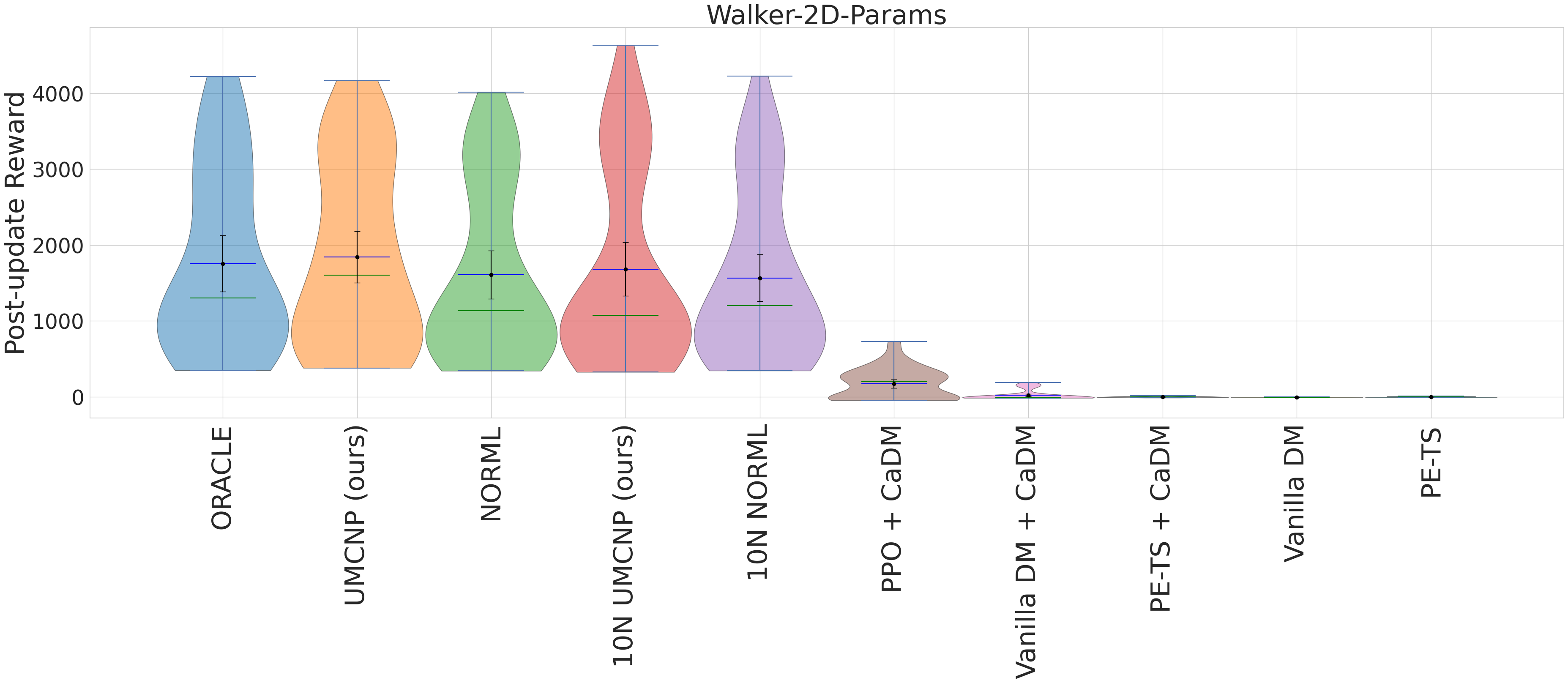}
\caption{Violin plots of post-update cumulative reward from unseen 10 meta-test tasks with 95\% Confidence Intervals (black). These plots combine vertically aligned kernel density plots that represent the data distribution, with markers clearly indicating the minimum, maximum, mean (blue), and median (green) values. 10N UMCNP (ours) and 10N NORML uses 10 test task transitions respectively.}
\vspace*{-3mm}
\label{fig:walkerexp}
\end{figure}

We use only 40 tasks for meta-training sampled from a uniform distribution and compare our method with the baselines on 10 test tasks similar to the meta-testing experiments with rewards\cite{rakelly2019efficient}. Violin plots in Fig. \ref{fig:walkerexp} show the post-update cumulative reward in meta-testing, where NORML, UMCNP use a single rollout, 10N NORML, 10N UMCNP use 10 transitions, and PE-TS CaDM\cite{lee2020context}, Vanilla CaDM\cite{lee2020context}, PPO-CaDM \cite{lee2020context} with normalization and Cross Entropy Method\cite{botev2013cross} for planning following \cite{lee2020context} use a history and future transition length of 10, and the ORACLE uses 25 rollouts for adaptation. Results in Table \ref{tab:cartwalk} indicate that UMCNP, not only improved sample efficiency but has the potential to improve meta-test adaptation performance for a larger range of unseen tasks. One reason for that can be the bias-variance trade-off in noisy real samples and generated samples. UMCNP network is trained on the meta-train dataset, hence it can generate less noisy and potentially more robust samples that are seen during training for the meta-policy which can make the adaptation more robust.
\begin{table}[!t]
\caption{Post-update cumulative Reward of UMCNP and the baselines \label{tab:cartwalk}}
\centering 
\begin{tabular}{l||c|c}
\hline  \hline $\mathbf{Cartpole}$ $\mathbf{Agent}$  & $\mathbf{mean}$&$\mathbf{median}$\\
\hline 5N NORML & 144.56 $\pm$ 48.72 & 45 \\
\hline 5N UMCNP (ours) & 424.13$\pm$ 39.79 & \textbf{499} \\
\hline PPO + CaDM & 295.40 $\pm$ 39.07 & 280 \\
\hline Vanilla + CaDM & 18.62 $\pm$ 2.32 & 16 \\
\hline PE-TS + CaDM & 52.82 $\pm$ 12.36 & 35 \\
\hline Vanilla DM & 78.22 $\pm$ 16.87 & 57 \\
\hline PE-TS & 82.87 $\pm$ 14.15 & 69 \\
\hline \textbf{50N UMCNP (ours)} & \textbf{427.02} $\pm$ 42.76 & \textbf{499} \\
\hline 50N NORML & 238.51 $\pm$ 55.59 & 174 \\
\hline \hline $\mathbf{Walker-2D}$ $\mathbf{Agent}$ & $\mathbf{mean}$&$\mathbf{median}$\\
\hline ORACLE & 1755.07 $\pm$ 370.89 & 1303.97 \\
\hline \textbf{UMCNP (ours)} & \textbf{1844.62} $\pm$ 338.80 & \textbf{1602.97} \\
\hline NORML & 1608.97 $\pm$ 316.83 & 1135.64 \\
\hline 10N UMCNP (ours) & 1684.56 $\pm$ 355.71 & 1072.68 \\
\hline 10N NORML & 1566.38 $\pm$ 308.84 & 1203.82 \\
\hline PPO + (Vanilla + CaDM) & 171.18 $\pm$ 57.00 & 199.05 \\
\hline Vanilla DM + CaDM & 22.31 $\pm$ 17.79 & -7.28 \\
\hline PE-TS + CaDM & -1.20 $\pm$ 1.86 & -2.47 \\
\hline Vanilla DM & -4.34 $\pm$ 0.37 & -4.47 \\
\hline PE-TS & -0.48 $\pm$ 1.45 & -2.42 \\
\hline
\end{tabular}
\vspace*{-3mm}
\end{table}

\section{Conclusion}
We propose Unsupervised Meta-Testing with Conditional Neural Processes, a new meta-RL with unsupervised meta-testing method that can generate useful rollouts for sample-efficient meta-policy adaptation. The agent performs without the need for rewards during meta-testing and learns without explicit task parameters, depending entirely on interaction data, during both meta-training and meta-testing. Our results indicate that our method's meta-test performance, when fine-tuned with generated data, shows significantly better performance in point and cartpole environments, and better performance in walker environments. Crucially, its performance is similar to the ground-truth method, which is allowed to interact with the environment significantly more. This sample efficiency across all experiments is achieved through the generation of samples from fewer number of transition tuples. 

In conclusion, we show that leveraging generated data for meta-updates not only improves sample efficiency but can also enhance post-update performance. Additionally, the successful application of CNPs to high-dimensional inputs, such as image data \cite{cnmp}, suggests promising avenues for future work. Specifically, applying our findings to high-dimensional sensorimotor spaces, such as manipulator robots equipped with RGB-D cameras, is an exciting direction for further exploration.
\vspace*{-3mm}
\appendix
Our model features a 128-dimensional representation and two hidden layers, each with 128 units, in both the encoder and decoder. It is trained with a batch size of 32 with  50 input tuples and 50 target query tuples for 500000 iterations with Adam optimizer \cite{kingma2014adam} with a learning rate of 0.0001. Additionally, the model utilizes the last 10\% of trajectories from 
$D^\text{train}$ which are gathered by the meta-policy throughout the meta-training phase.
\vspace*{-3mm}
\section*{ACKNOWLEDGMENT}
This work was funded by TÜBİTAK BİDEB 2224-A Grant Program for Participation in Scientific Meetings Abroad. Emre Ugur was partially supported by EU-funded INVERSE poject (no.101136067). Numerical calculations were partially performed at TUBITAK ULAKBIM TRUBA resources. We thank the anonymous reviewers, editors for their valuable feedback and constructive suggestions.
\vspace*{-3mm}
\bibliographystyle{IEEEtran}  
\bibliography{IEEEabrv,references}

\end{document}